\documentclass[10pt,twocolumn,letterpaper]{article}

\usepackage{iccv}

\usepackage{epstopdf}
\usepackage{graphicx}    
\usepackage{subcaption}
\usepackage{float}
\usepackage[justification=raggedright]{caption}	
\usepackage{lscape}                                         

\graphicspath{{figures}, {example}}

\usepackage{multirow}
\usepackage{tabularx, booktabs}

\usepackage{bm}                          
\usepackage{epsfig}                      
\usepackage{graphicx}                  
\usepackage{times}
\usepackage{amssymb,amsmath}   
\usepackage{verbatim}
\usepackage{array}

\usepackage{units}
\usepackage{color}

\usepackage{url}  
\usepackage[pagebackref=true,breaklinks=true,letterpaper=true,colorlinks,bookmarks=false]{hyperref}

\usepackage{cite}

\def\etal{et~al.~}			  
\def\eg{e.g.,~}                
\def\ie{i.e.,~}                  

\DeclareMathOperator*{\argmin}{\arg\!\min}

\newlength\paramargin
\newlength\figmargin
\newlength\secmargin

\long\def\ignorethis#1{}

\newcommand{\tb}[1]{\textbf{#1}}

\newcolumntype{C}{>{\centering\arraybackslash}p{5ex}}

\iccvfinalcopy 

\def\iccvPaperID{581} 

\ificcvfinal\fi
\begin{document}

\title{Optimizing Region Selection for Weakly Supervised Object Detection}

\author{Wenhui Jiang{\small $~^{1}$}, Thuyen Ngo{\small $~^{2}$}, B.S. Manjunath{\small $~^{2}$}, Zhicheng Zhao{\small $~^{1}$}, Fei Su{\small $~^{1}$} \\
\\
{\small $~^{1}$} Beijing University of Posts and Telecommunications, Beijing, China\\
{\small $~^{2}$} University of California, Santa Barbara, CA, USA\\
}

\maketitle

\begin{abstract}
   Training object detectors with only image-level annotations is very challenging because the target objects are often surrounded by a large number of background clutters. Many existing approaches tackle this problem through object proposal mining. However, the collected positive regions are either low in precision or lack of diversity, and the strategy of collecting negative regions is not carefully designed, neither. Moreover, training is often slow because region selection and object detector training are processed separately. In this context, the primary contribution of this work is to improve weakly supervised detection with an optimized region selection strategy. The proposed method collects purified positive training regions by progressively removing easy background clutters, and selects discriminative negative regions by mining class-specific hard samples. This region selection procedure is further integrated into a CNN-based weakly supervised detection (WSD) framework, and can be performed in each stochastic gradient descent mini-batch during training. Therefore, the entire model can be trained end-to-end efficiently. Extensive evaluation results on PASCAL VOC 2007, VOC 2010 and VOC 2012 datasets are presented which demonstrate that the proposed method effectively improves WSD. 
\end{abstract}

\section{Introduction}

Object detection aims at predicting the category as well as the bounding box locations for each object instance. Recent breakthroughs in object detection~\cite{Girshick2015,Ren2015,Dai2016,Redmon2016You,liu2016ssd} are driven by supervised approaches with deep convolutional neural networks (CNNs). However, strong supervision requires a large number of bounding box annotations, which are relatively hard to obtain~\cite{Su2012,Papadopoulos2014,Uijlings2016}. In contrast, weakly-supervised detection (WSD) assumes that only image-level labels indicating the presence or absence of an object category are available for training. 

\begin{figure}[!t]
\centering
\epsfig{file=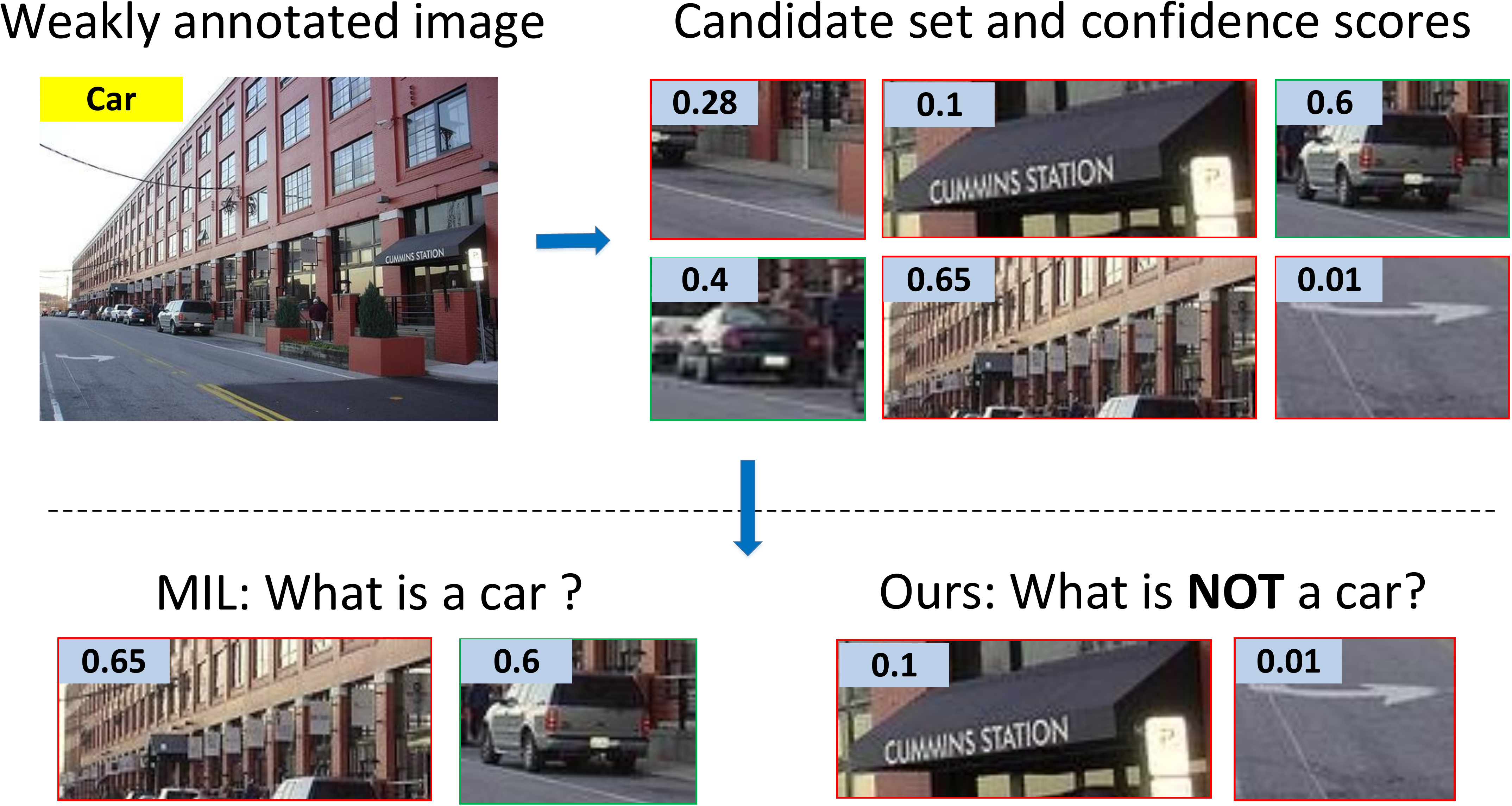,width=\linewidth}
\caption{Comparison of our work with existing multiple instance learning (MIL) based approaches on positive region mining. Green boxes indicate true positive regions while the red ones represent backgrounds. MIL-based models select positive regions from a noisy candidate set directly, which usually results in low accuracy and diversity (bottom-left). In contrast, we progressively remove clear background regions from the candidate set (bottom-right), which reduces object-background ambiguity effectively and helps object localization easier and accurate. On VOC2007, our strategy improves the detection accuracy by 5.8\% over the state-of-the-art MIL-based model~\cite{Wang2015}.}
\label{fig:1}
\end{figure}

Most existing methods formulate the weakly supervised detection task as a multiple instance learning (MIL) problem~\cite{Song2014,Bilen2015,Wang2015,Cinbis2016,Ren2016}. Under MIL paradigm, learning usually alternates between two steps. (a) Estimating CNN-based object detectors based on a fixed set of training regions. (b) Updating the training set using the learned object detectors. Although existing methods have shown promising results, these methods have two major drawbacks.
\begin{enumerate}
\item The collected positive sets are either noisy or lack of diversity (see Figure~\ref{fig:1} for an example). As a result, the performance of the learned object detectors is limited. 

\item No model updates are made in step (b)---the CNN-based object detectors are frozen for at least one epoch for updating on the entire training set. This largely slows down the training process. 
\end{enumerate}

\begin{figure*}[!t]
\centering
\epsfig{file=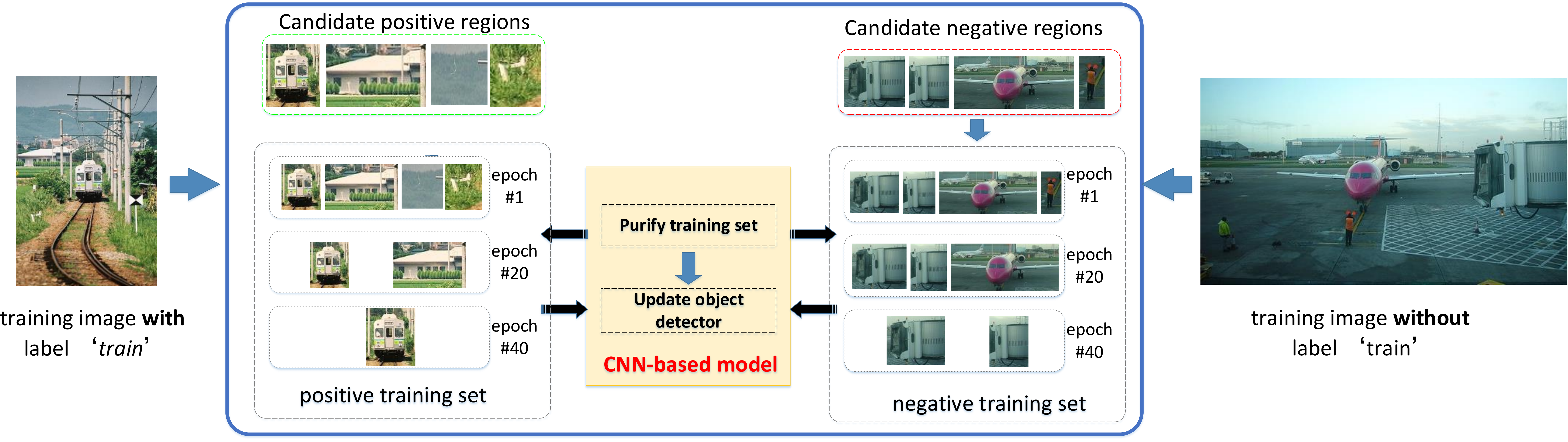, width=\linewidth}
\caption{A brief illustration of the relevant region selection precess. At the beginning, for a given object category (\eg``train''), the positive training set starts from all the regions of the images with positive labels, and the negative training set begins with  all the regions of the images with negative labels. During the training iteration, obvious negative samples inside positive set are filtered out, leading to a purified positive set. At the same time, the easy negative samples inside negative set are also removed, leading to a discriminative negative set.}
\label{fig:2}
\end{figure*}

In this paper, we attempt to address two major questions. First, how to select positive and negative regions for training CNN-based object detectors under weak supervision. Ideally, the selected regions should be accurate and discriminative. Second, how region selection can be combined with the learning of object detectors so that both parts can be jointly optimized during training. 


Towards the first goal, we propose an optimized region selection strategy. It consists of positive and negative region selection. To select positive regions, \emph{our key observation is that it is hard to identify all true positives directly for object detectors.} Essentially, the main difficulty arises from the large ambiguity between small target objects and large background clutters. In contrast, as shown in Figure~\ref{fig:1}, \emph{it is relatively easy to find out most of the background regions for most models.} Hence, we progressively update the positive set by filtering out easy backgrounds while training the object detectors. As the number of background regions decreases gradually, object-background ambiguity can be reduced effectively and localizing true positive regions becomes easier. We also investigate a class-specific hard negative selection strategy to improve the discrimination of negative training set. Unlike positive labels, negative labels provide strong supervision, \ie all of the regions are negative samples if the image-level label is negative. As has been revealed in previous studies~\cite{Shrivastava2016,Felzenszwalb2010}, the ``hard'' samples (which are wrongly determined by the detector) are more discriminative compared with the ``easy'' ones. Therefore, we select hard regions inside negatively-labeled images as negative samples. The negative samples are class-specific, hence they can carry more discriminative information. This is illustrated in Figure~\ref{fig:2}. 

To address the second issue, we incorporate both region selection and object detector learning into a unified CNN framework. Specifically, in each  stochastic gradient descent (SGD) mini-batch, the region selection module samples the class-specific training regions through the forward process, then the object detectors are updated on top of them. Therefore, the proposed model can be updated as frequently as the networks without region selection~\cite{Bilen2016}. 


Our proposed approach is simple to implement. We evaluate and compare the performance of our method on three challenging datasets, the PASCAL VOC 2007, VOC 2010 and VOC 2012~\cite{Everingham2014}. Our model achieves the detection average precision (mAP) of 37.4\% on VOC 2007, which is 5.6\% higher than the baseline without region selection~\cite{Bilen2016}, and 5.8\% higher than the best MIL-based model~\cite{Wang2015}. On VOC 2010 and VOC 2012, we obtain the precision of 36.0\% and 33.6\%, which significantly outperform other state-of-the-art methods.

Summarizing, the main contributions of our work are:
\begin{enumerate}
  \item We propose a progressive pruning strategy for region selection. This algorithm leads to purified positive sets and discriminative negative sets, both of which improved detection performance.
  \item Our model addresses both object detector learning and region selection within a unified CNN network, which allows for end-to-end training. Although region selection has been discussed in several works~\cite{Andrews2002,Siva2012,Ren2016} before, we apply it to build a novel CNN network for weakly supervised detection, which is, to the best of our knowledge, unique to our work.
\end{enumerate}


\section{Related work}
\subsection{Multiple Instance Learning}
There have been a number of recent works addressing weakly-supervised detection (WSD) through the Multiple Instance Learning (MIL) paradigm~\cite{Andrews2002,Song2014a,Bilen2015,Wang2015,Ren2016,Cinbis2016}. Several methods focus on selecting precise positive regions and discriminative negative regions to improve WSD. For example, Andrews \etal\cite{Andrews2002} pick out the most confident region from each positive (or negative) image as the positive (or negative) training sample. Siva \etal\cite{Siva2012} localize the positive regions which have maximal distances to their nearest neighbors within negative images. Ren \etal\cite{Ren2016} propose a bag-splitting algorithm that iteratively includes hard negative regions from positively-labeled images into negative training set, and randomly selects negative samples from the set. 


Our work is inspired by previous work on selecting training regions, but different in three aspects. First, our model focus on a sparse collection of positive regions, which improves the diversity of the positive training samples and avoids the impact from noisy backgrounds as well. Second, instead of selecting easiest negative region~\cite{Andrews2002} or randomly~\cite{Ren2016}, we collect class-specific hard samples as negative regions which carry more discriminative information. Third, different from most MIL-based methods which process region selection in an offline manner, our proposed region selection strategy can be processed in each SGD mini-batch, which makes the training efficient.


\subsection{CNN for weakly supervised detection}
As another line of research, Oquab \etal\cite{Oquab2014,Oquab2015}, Zhou \etal\cite{Zhou2015,Zhou2016} and Bency~\cite{Bency2016} show that CNN for image classification automatically learns object activation maps, where the target objects can be coarsely localized. In these works, the activation maps can be learned end-to-end from image-level labels. However, they require a separate post-processing step to obtain the final localization. 

The most similar approach to ours is the weakly supervised deep detection network (WSDDN) by Bilen \etal\cite{Bilen2016}. In \cite{Bilen2016}, a two-stream CNN network is proposed---one stream for region classification and the other stream for detection. However, WSDDN has two major drawbacks. On one hand, it attempts to train object detectors from the whole noisy candidate region set. On the other hand, it tends to capture only the simplest negative samples. While our network architecture is similar to WSDDN~\cite{Bilen2016}, the region selection module encourages our network to focus a sparse collection of accurate positive regions and discriminative negative regions, which improves the distinction of the learned object detectors. This is explained in greater detail and verified experimentally in section~\ref{selection}. In addition, such improvement does not involve extra parameters compared with WSDDN~\cite{Bilen2016}. 


\subsection{Online batch selection}
There are several methods~\cite{Loshchilov2016,Shrivastava2016} that select hard training samples for training CNNs under strong supervision. The hard samples are chosen based on losses. Our work is motivated by~\cite{Loshchilov2016,Shrivastava2016}. However, under weak supervision, the relevant regions can not be simply defined based on losses, therefore it is not straightforward to adapt existing methods to WSD. Moreover, existing methods~\cite{Loshchilov2016,Shrivastava2016} only select class-independent hard samples. In contrast, in our work, the selected training samples are class-specific, which carries more discriminative information.

\begin{figure*}[!t]
\centering
\epsfig{file=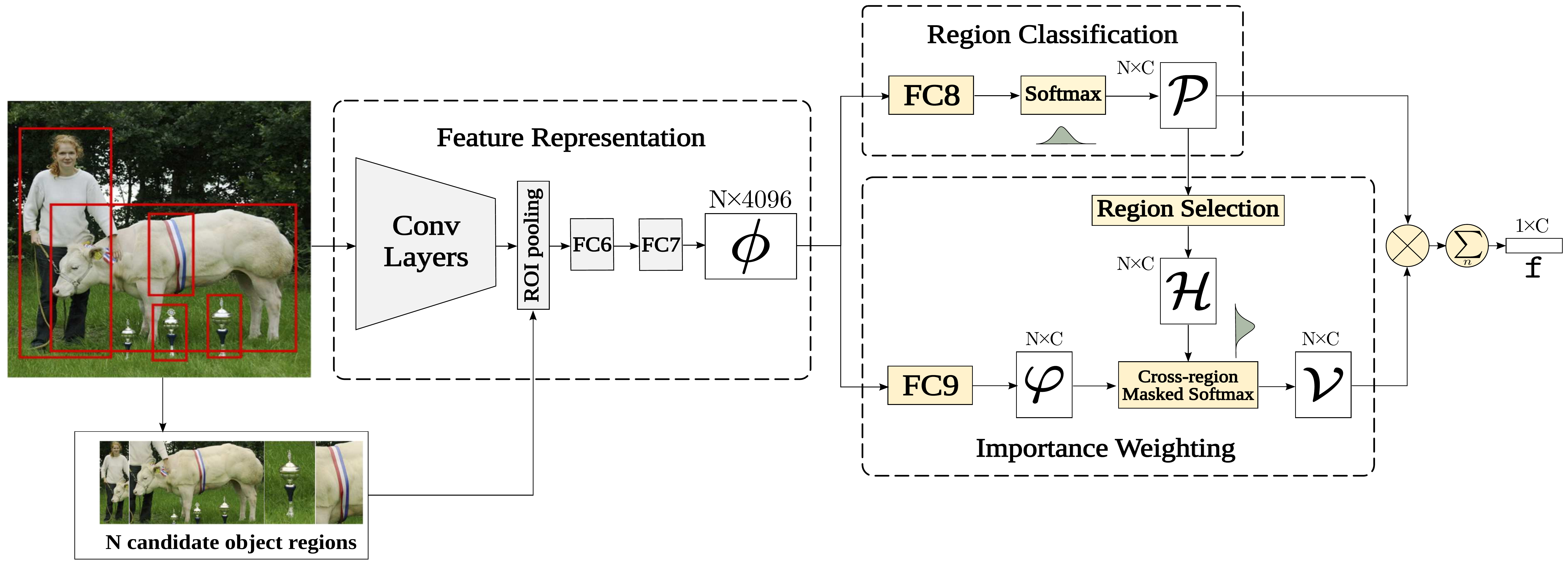,width=\linewidth}
\caption{Overall architecture of our proposed model. It mainly consists of three modules: 1) a feature representation module which extracts a deep feature vector $\bm{\phi}$ for each candidate object region, 2) a region classification module which estimates class probability vector $\textbf{p}$ for each region individually, and 3) an importance weighting module, which selects relevant training regions and estimates sample importance vector $\textbf{v}$ for each region. Please see Section \ref{formulation} and \ref{network} for detailed illustrations.}
\label{fig:3}
\end{figure*}

\section{WSD with region selection} \label{formulation}
In this section, we briefly describe our method for weakly supervised detection. The model is designed to automatically select relevant training regions in each SGD mini-batch, and simultaneously performs end-to-end learning of a deep object detector from the selected regions.

Given $K$ training images $\{I^k\}_{k=1}^{K}$, each image $I^k$ has a set of candidate object regions $B^k=\{b_{i}^{k}\}_{i=1}^{N^k}$. Let $\textbf{y}^{k}=[y_{1}^{k},...,y_{C}^{k}]$ represents its image-level label vector, where $C$ is the total number of object categories and $y_{c}^{k}\in \{+1, 0\}$ denotes whether this image contains the $c$-th object category. Based on the assumption of Multiple Instance Learning, if $y_{c}^{k} = +1$, then at least one region $b_{i}^k \in B^k$ is a positive sample for the $c$-th category. Otherwise, all of the regions in $B^{k}$ are negative samples.

Let $\textbf{p}_{i}^k=[p_{i,1}^{k},...,p_{i,C}^{k}]$ denote the estimated class probability vector for candidate object region $b_{i}^k$. Denote $\{\textbf{v}_{i}^k\}_{i=1}^{N^{k}}$ as the weighting vectors for all regions, where $\textbf{v}_{i}^k=[v_{i,1}^{k},...,v_{i,C}^{k}]$ and $v_{i,c}^{k} \in [0,1]$ represents the sample importance of region $b_{i}^k$ for training the $c$-th object category. Different from $\textbf{p}_{i}^k$ which is only determined by each region $b_{i}^k$ individually, $\textbf{v}_{i}^k$ determines the sample importance by comparing among all the regions inside the image $I^{k}$. The correct and discriminative regions will be selected and learned to have high values to emphasize their sample importance. The rest regions will result in $v_{i,c}^{k} = 0$. In this work, we model $\textbf{v}_{i}^k$ and $\textbf{p}_{i}^k$ with an importance weighting module and a object classification module, respectively. Assume $\textbf{v}_{i}^k$ and $\textbf{p}_{i}^k$ are determined by parameter vector $\bm{\theta}$. We can define a per-category per-image cross entropy loss $\mathcal{L}$:
\begin{equation}\label{eq1}
\begin{split}
\mathcal{L}[y_{c}^{k}, f(B^{k};\bm{\theta})] &= - y_{c}^{k}\log(f(B^{k};\bm{\theta})) \\
& - (1-y_{c}^{k})\log(1-f(B^{k};\bm{\theta}))
\end{split}
\end{equation}

\noindent where $f(B^{k};\bm{\theta})$ is the aggregation function~\cite{Pinheiro2015,Durand2016} which aggregates the region-level class probabilities into image-level annotation scores:

\begin{equation}\label{eq2}
\begin{split}
& f(B^{k};\bm{\theta}) = \sum_{i=1}^{N^{k}}(v_{i,c}^{k}\cdot p_{i,c}^{k}) \\
& \text{s.t.  }\sum_{i=1}^{N^{k}}v_{i,c}^{k}= 1, \text{  }\sum_{c=1}^{C}p_{i,c}^{k}= 1
\end{split}
\end{equation}

\noindent It is worth mentioning that the constraint on $\textbf{v}_{i}^k$ is applied across all the regions, which is different from that of $\textbf{p}_{i}^k$. Such constraint is designed to allocate higher weights on only important regions.

In our model, the final goal is to learn the model parameter vector $\bm{\theta}$ according to the summed loss:

\begin{equation}\label{eq3}
\bm{\theta} = \underset{\bm{\theta}}{\argmin}\sum_{c=1}^{C}\sum_{k=1}^{K}\mathcal{L}[y_{c}^{k}, f(B^{k};\bm{\theta})]
\end{equation}

We propose a novel network to model Eqs. (\ref{eq1})-(\ref{eq3}), which will be explained in detail in the following section.

%

\section{Network architecture} \label{network}
\subsection{Overview}
Figure~\ref{fig:3} shows the architecture of the proposed model. Our model is built based on weakly supervised deep detection network (WSDDN) architecture~\cite{Bilen2016}. It takes an image $I^{k}$ and a set of candidate regions $B^{k}=\{b_{i}^k\}_{i=1}^{N^{k}}$ as inputs (for simplicity, we drop the superscript $k$ in the rest of the paper). It mainly consists of three modules: 1) a feature representation module, 2) a region classification module, 3) an importance weighting module. The feature representation module extracts fixed-size deep descriptor $\bm{\phi}_{i}$ for each candidate region $b_{i}$. The region classification module maps $\bm{\phi}_{i}$ to a class probability vector $\textbf{p}_{i}$ for each region individually. The importance weighting module selects relevant positive and negative regions among all the regions, and outputs a weighting vector $\textbf{v}_{i}$ which reflects relative importance of regions for training the entire network. The learning of the three modules is supervised by image-level annotation $\textbf{y}$ only. In the following subsections, we will explain these three modules in more details.

\subsection{Feature representation}
This module is built based on Fast RCNN architecture~\cite{Girshick2015}. It first takes an image $I$ of arbitrary size as input, extracting image feature maps $\textbf{G}$ using a stack of convolutional layers (\eg five convolutional layers as in AlexNet~\cite{Krizhevsky2012}). Then, given each candidate region $b_{i}$, a fixed-size region-level feature map is extracted through ROIPooling layer~\cite{Girshick2015} on $\textbf{G}$. Two fully connected layers (FC6 and FC7) are followed to output a fixed-length feature vector $\bm{\phi_{i}}$.

\subsection{Region classification}
This module is designed to classify $b_{i}$ into object categories. Towards this goal, we apply an additional fully connected layer (FC8) that maps $\bm{\phi}_{i}$ to a $C$ dimensional output $\bm{\phi}_{8i}$. As each region could represent at most one object, we apply conventional softmax operator to normalize $\bm{\phi}_{8i}$ to a class probability vector $\textbf{p}_{i}$.

\subsection{Importance weighting}
This is the most important part of our model. It mainly contains two components, one is to select class-specific training regions, the other one is to impose soft sample importance on the selected regions. The module outputs weighting vectors $\{\textbf{v}_{i}\}_{i=1}^{N}$, which reflect the relative sample importance of each region in image $I$.

\noindent\textbf{Region selection}: For each object category $c$, we choose positive regions from images with positive label, and negative regions from images with negative labels. We consider the following circumstances:

\begin{enumerate}
  \item When $y_{c} = +1$, if the estimated score $p_{i,c}$ is low, then it is very likely that $b_{i}$ comes from the background, and therefore should be removed from the positive training set ($v_{i,c} = 0$). The rest regions are candidate true positives and are taken as positive training set ($v_{i,c} > 0$). 
  \item When $y_{c} = 0$, if the estimated score $p_{i,c}$ is high, then $b_{i}$ is wrongly classified, and is therefore selected as a negative region ($v_{i,c} > 0$). This coincides with the idea for hard negative mining~\cite{Felzenszwalb2010}, which indicates that "hard samples" are more discriminative under strong supervision.
\end{enumerate}

In summary, only the regions with high probability scores $p_{i,c}$ are selected as training samples. Therefore, for each object category, we rank the regions based on probability scores $p_{i,c}$, and pick up the top $M_{pc}$ or $M_{nc}$ highest scoring regions as positive or negative regions according to the image-level labels. For simplicity, we assume the value of $M_{pc}$ and $M_{nc}$ are independent of object categories. Therefore, we denote $M_{pc}$ and $M_{nc}$ as $M_{p}$ and $M_{n}$, respectively. Formally, region selection can be processed according to:

\begin{equation}\label{eq4}
\begin{split}
  & \{h_{i,c}\}_{i=1}^{N} = \underset{\textbf{h}}{\mathrm{argmax}}\sum_{i=1}^{N}(h_{i,c}\cdot p_{i,c}) \\
  & \text{s.t.  }\sum_{i=1}^{N}h_{i,c}= y_{c}M_{p} + (1-y_{c})M_{n}
\end{split}
\end{equation}

\noindent where $\textbf{h}_{i}=[h_{i,1},...,h_{i,C}]$ represents the selection indicator vector for region $b_{i}$, $h_{i,c}\in \{1, 0\}$ indicates the selection of $b_{i}$ for training the $c$-th object category or not.

For different object categories, the training samples might be different. Such class-specific region selection strategy carries more discriminative information, because each class can select the most relevant regions against the other classes.


\noindent\textbf{Soft importance weighting}: We apply a cross-region soft weighting strategy to impose different sample importance on the selected regions. Specifically, we append another fully connected layer (FC9) which maps $\bm{\phi}_{i}$ to a $C$ dimensional output $\bm{\varphi}_{i}$. Then, $\bm{\varphi}_{i}$ is connected to a cross-region masked softmax layer with $\textbf{h}_{i}$, which is defined as:

\begin{equation}\label{eq5}
v_{i,c} = \frac{h_{i,c}\exp(\varphi_{i,c})}{\sum_{j=1}^{N}h_{j,c}\exp(\varphi_{j,c})}
\end{equation}

As we can see, $v_{i,c}$ compares the relative sample importance of each region for training the $c$-th object category. If a certain region is masked by $h_{i,c}$, $v_{i,c}$ outputs 0. This region selection strategy proceeds within a single CNN forward operation. During backward operation, only the selected regions contribute to gradient propagation. Therefore, the whole model can be updated as frequently as the same network without region selection.

\subsection{Progressive pruning}
\begin{figure}[!t]
\centering
\epsfig{file=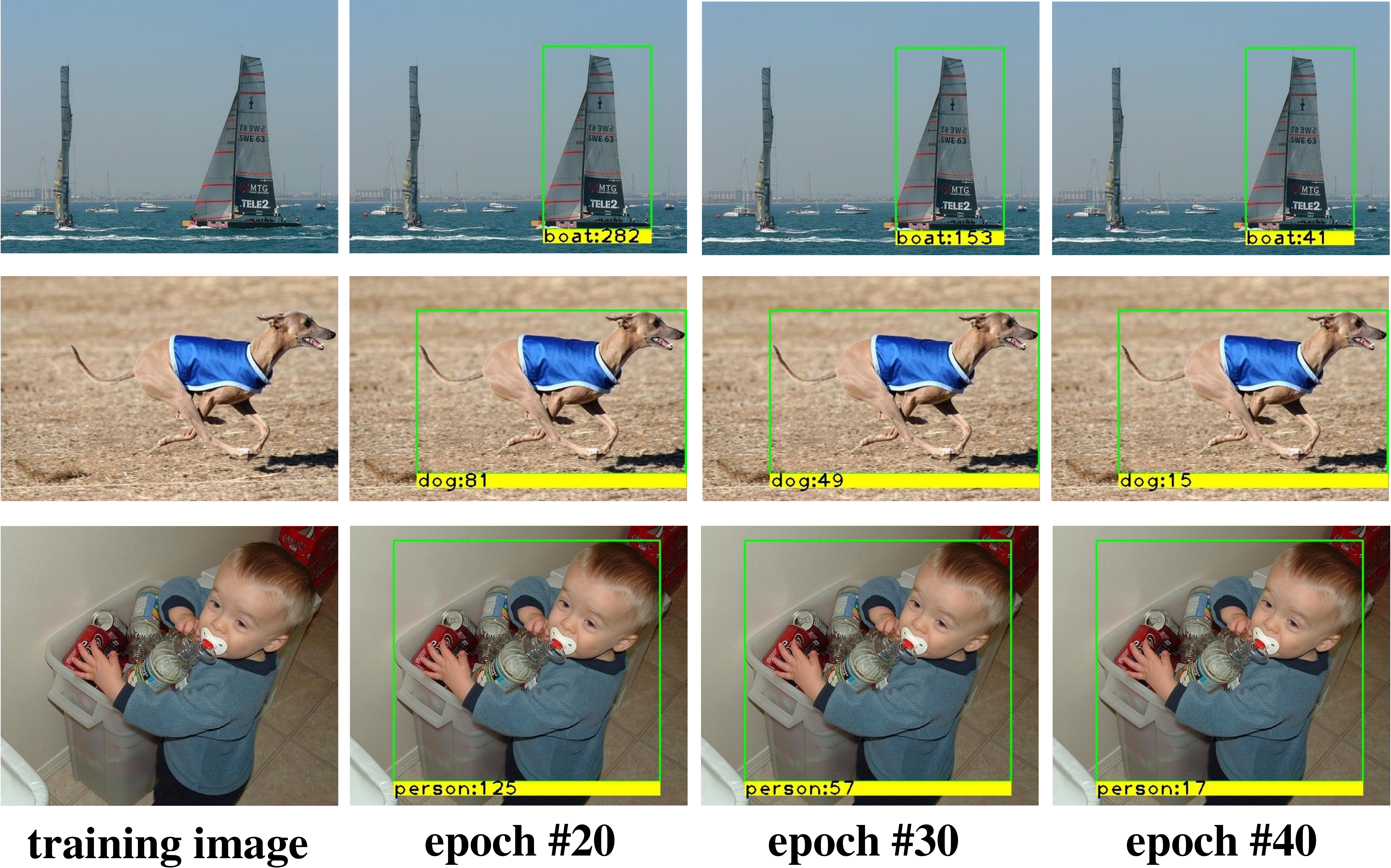, width=\linewidth}
\caption{Update of the rankings of sampled positive regions during training. In each yellow rectangle, we list the ranking of the labelled region among all proposals inside the given image based on the class probability score.}
\label{fig:progressive}
\end{figure}

The parameter $M_{p}$ controls the pace at which the model learns from positive samples. In this paper, we propose to adjust $M_{p}$ dynamically as the training processes. Intuitively, when the learned detectors are weak, we set $M_{p}$ to a large number so that high recall rates are retained. As the detectors grow mature, more positive regions are ranked in the top (see Figure \ref{fig:progressive} for visual examples), we gradually decrease $M_{p}$ to improve the precision of the positive set. Specifically, $M_{p}$ is set to $N$(the total number of candidate regions inside the image) when the training begins. After 20 epoches, we set $M_{p}$ to 1024, and progressively reduce $M_{p}$ by half every $N_{e}$ epoches until $M_{p}$ reaches a pre-defined threshold $M_{pt}$. $N_{e}$ is obtained according to:

\begin{equation}\label{eq6}
N_{e} = \frac{20}{\log_{2}(1024/M_{pt}) + 1}
\end{equation}

Since negative image labels provide strong supervision, we keep $M_{n}$ fixed during the training process. In practice, we set both $M_{pt}$ and $M_{n}$ to 128, which is a typical setting for mini-batch selection.


\section{Experiments}
\subsection{Datasets}
We evaluate our model on three large image datasets, the PASCAL VOC 2007, VOC 2010 and VOC 2012. The VOC 2007 dataset has labels for 20 object categories and contains 2501 images for training, 2510 images for validation and 4952 images for testing. VOC 2010 and VOC 2012 share the same class labels. VOC 2010 includes 4998 images for training, 5105 images for validation and 9637 images for testing. VOC 2012 contains 5717 training images, 5823 validation images and 10991 test images. These datasets contain both image-level labels and object location annotations. For weak supervision, we only utilize the image-level labels for training. Following~\cite{Wang2015,Bency2016,Cinbis2016,Bilen2016}, we use both train and val splits as the training set and test split as our test set for VOC 2007 and VOC 2010. For VOC 2012, we use train split as the training set and val split as the test set.

\subsection{Evaluation metrics}
We use two metrics to evaluate localization performance. First, we quantify localization performance in the training set with the Correct Localization (CorLoc) measure~\cite{Deselaers2012}. CorLoc is the percentage of images in which the bounding-box returned by the algorithm correctly localizes an object of the target class. It reflects the top-1 accuracy of positive region selection strategy. Second, we measure the performance of object detectors using mean average precision (mAP) in the test set, as standard in PASCAL VOC. For both metrics, we consider that a bounding box is correct if it has an intersection-over-union (IoU) ratio of at least 0.5 with a ground-truth object instance annotation.

\subsection{Implementation details} \label{Implementation}
Our network is built with caffe~\cite{jia2014caffe}. We adopt AlexNet~\cite{Krizhevsky2012} as the backbone architecture. It is pre-trained on Image{N}et~\cite{Russakovsky2015} to initialize the convolutional layers and the two fully connected layers (FC6 and FC7). Rest of the layers are initialized randomly as in~\cite{Girshick2015}.

We extract candidate object regions with EdgeBoxes (EB)~\cite{edgeboxes14} for each image. We fine-tune the network on the target datasets. Each mini-batch contains all the ROIs from one image. We adopt multi-scale training. Specifically, the longer side of images is resized to a random scale $s$ ($s\in \{480,576,688,864,1200\}$) while the aspect ratio is kept unchanged. We also apply random horizontal flips to the images for data augmentation. The experiments are run for 40 epoches. 

At test time we take $v_{i,c}\cdot p_{i,c}$ as the final class confidence score for region $b_{i}$. As with~\cite{Bilen2016}, we average the outputs of 10 images (\ie the 5 scales as in training and their flips). The results are post-processed by bounding box voting~\cite{gidaris2015object} and non-maximum suppression (NMS) using a threshold of 0.6 IoU.


\subsection{The impact of region selection}\label{selection}
\noindent\textbf{Settings}: To analyze the effects of region selection, we train the system with different values of $M_{p}$ and $M_{n}$ using VOC 2007 training set and measure the performance on 500 holdout images randomly selected from the validation set. 

\noindent\textbf{Baseline}: We remove the region selection module as the baseline. This is equivalent to setting $M_{p}$ and $M_{n}$ to $N$ for the entire training process. The baseline model achieves a mAP of 29.8\% on the holdout validation set. 

\begin{figure}
\centering
\includegraphics[width=\linewidth]{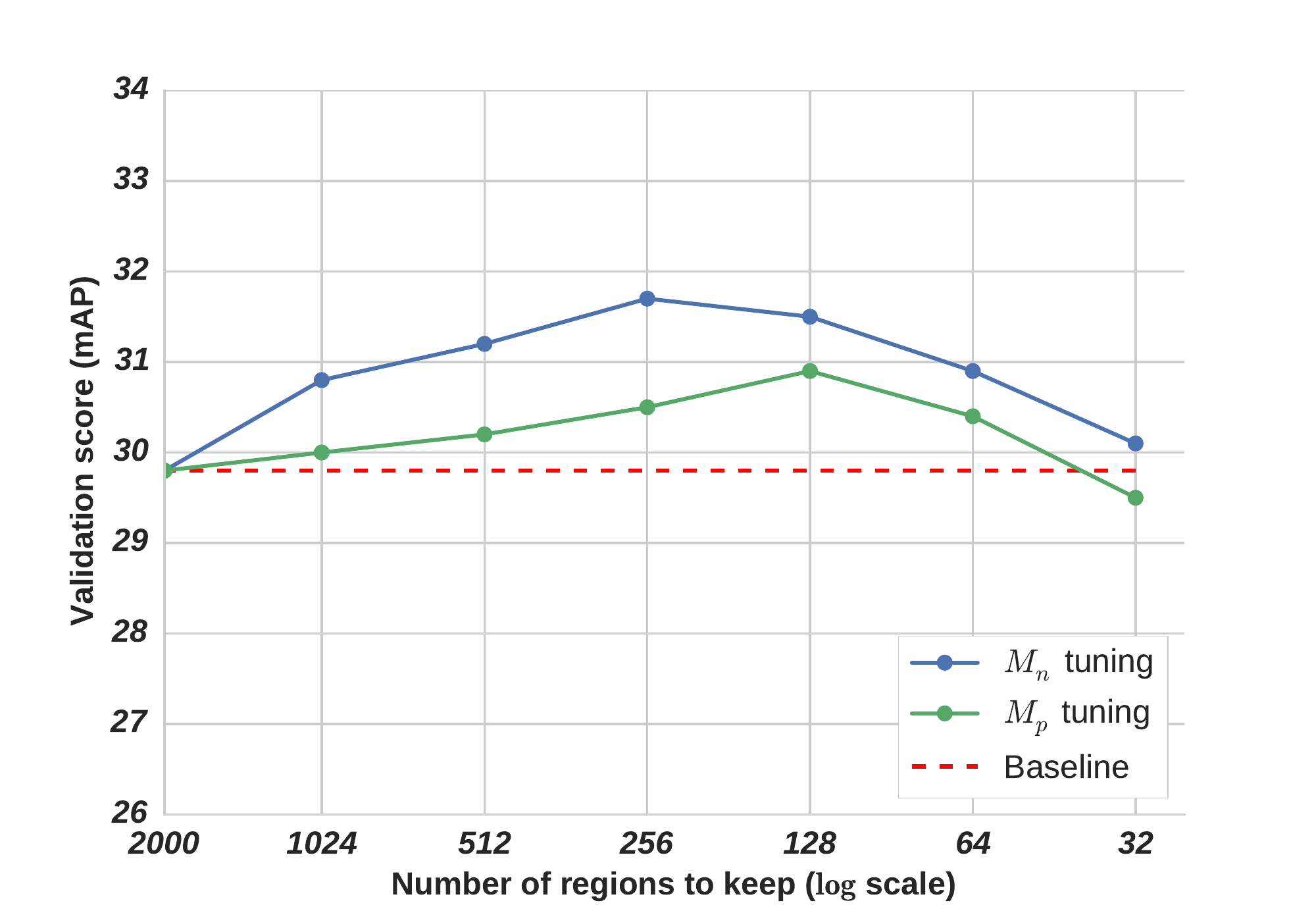}
\caption{Detection performance on VOC 2007 holdout images with different values of $M_p$ (number of selected positive regions) and $M_n$ (number of selected negative regions). Green curve: varying $M_p$ while fixing $M_n$. Blue curve:  varying $M_n$ while fixing $M_p$. Red dash line:  baseline (without region selection). See section \ref{selection} for more details.}
\label{fig:4}
\end{figure}

\noindent\textbf{Observation}: First of all, we demonstrate the importance of positive region selection. Toward this goal, we fix $M_{n}=N$ (which disable negative region selection), and gradually reduce $M_{p}$ from $N$ to 32 to see how positive region selection impacts WSD. As is shown in Figure~\ref{fig:4} (in green curve), when $M_{p}$ varies from $N$ to 32, the performance first monotonically increases, which demonstrates positive region selection helps improve object detection performance. When $M_{p}=128$, the mAP reaches the peak 30.9\%. Then the performance drops to 29.5\%. Second, we show the efficacy of negative region selection. In this experiment, we fix $M_{p}=N$ and decrease $M_{n}$ from $N$ to 32. As is shown in Figure~\ref{fig:4} (in blue curve), the mAP monotonically increases until it reaches the peak, then the performance drops again when $M_{n}$ continues decreasing. Setting $M_{n}$ to 256 leads to close-to-optimal result.

\noindent\textbf{Analysis}: It is important to discuss why relevant region selection helps improve the performance of object detection. We first look into the baseline model. As we can see from Eqs. (\ref{eq1}) - (\ref{eq2}), when $y_{c}=1$, regions with high classification score $p_{i,c}$ naturally encourage $v_{i,c}$ to be high as well, while regions with low $p_{i,c}$ will result in low $v_{i,c}$. With VOC 2007 train set, when the candidate regions are sorted by $p_{i,c}$, we observe the long-tail phenomenon --- 84.17\% of the importance weights are from the top-128 regions. We believe that suppressing the weights from regions ranking 128 and beyond can benefit WSD in two aspects: 1) more supervision flows towards true positive regions (most of them rank top-128), 2) less supervision mis-flows towards false negative regions (ranking 128 and beyond). As the weights occupied from 512 and beyond are ignorable, when $M_{p}$ drops from $N$ to 512, the performance boost is very limited. When we further reduce $M_{p}$ to 128, the mAP improves significantly. When $M_{p}$ comes to 32, as more true positive regions are filtered out (which leads to average recall rates decrease), the performance drops drastically.

\begin{figure}[!t]
\centering
\epsfig{file=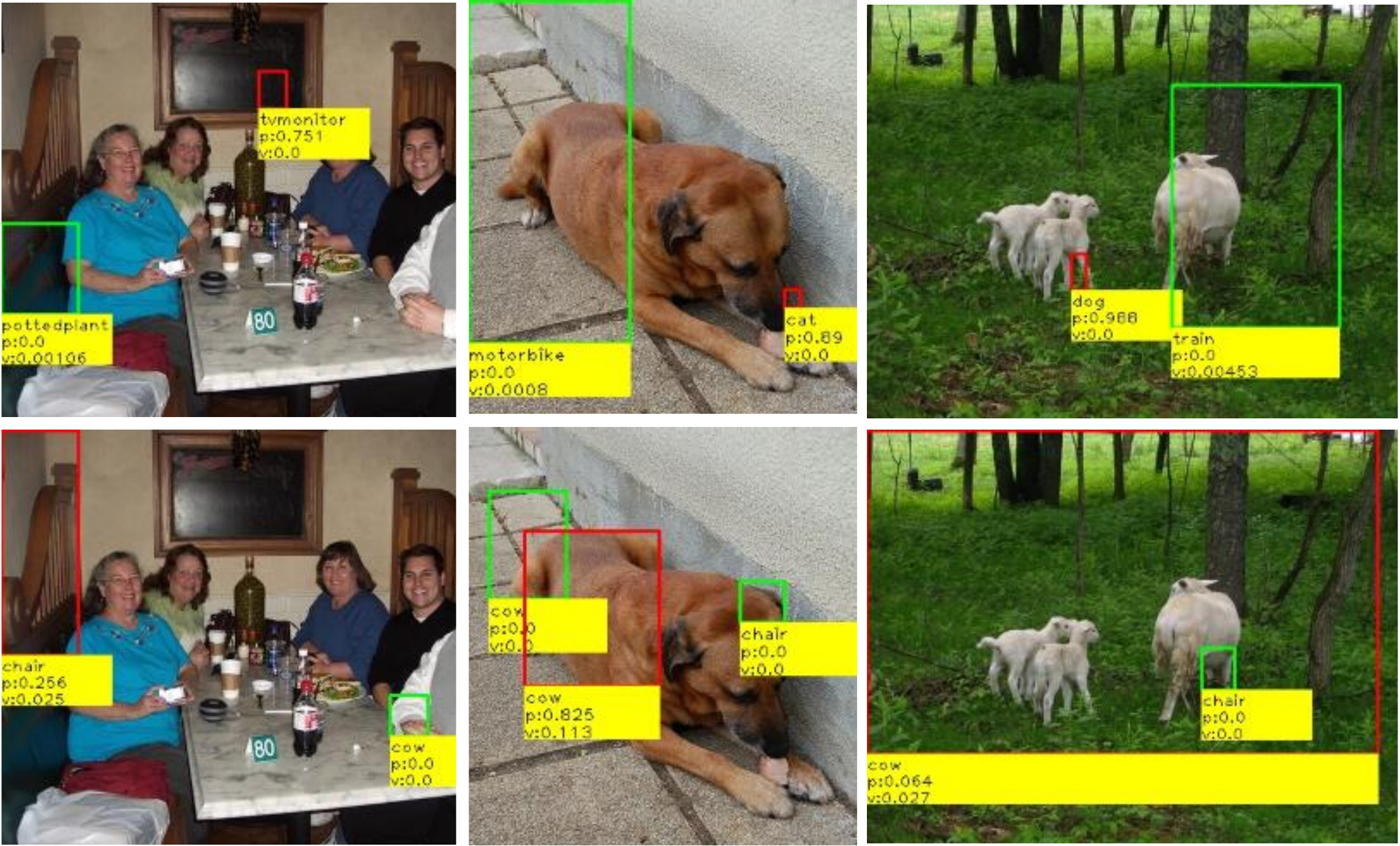, width=\linewidth}
\caption{Visualization of the negative regions. The green bounding boxes represent true negatives while the red boxes depict false negatives. $p$ represents the classification score and $v$ represents the weighting score. Top row shows the results from the baseline (without negative region selection), and the bottom row shows the results from our proposed model (with negative region selection).}
\label{fig:5}
\end{figure}

We also investigate the weight distribution when $y_{c}=0$. We observe that only 10.08\% weights come from the top-128 ranked regions. This is not surprising since the baseline model treats the supervision from both positive label and negative label equally, therefore it puts more weights on easy negative regions instead of hard ones. The negative region selection benefits WSD by re-assigning more weights on hard regions, and suppressing the weights on those correctly identified negative regions at the same time. For a better illustration, we visualize the ``hardest negative regions'' identified by our model (with negative region selection) and the baseline (without negative region selection) in Figure~\ref{fig:5}. In the examples, all the red boxes are false negatives with high probability scores, and green boxes are true negatives. However, for the baseline method (top row), the false negatives are assigned close-to-zero weights, indicating they are not important as negative regions for training object detectors. On the contrary, the true negatives are assigned high weights. In comparison, with negative region selection (bottom row), the hard negative regions are assigned higher weights. As a result, when we reduce $M_{n}$ from $N$ to 256, the performance improves consistently. When we further decrease $M_{n}$ to 32, the detection performance drops. This is probably due to the decrease in the diversity of negative regions within a mini-batch.

\subsection{Compare with other WSDDN variants}
Our network is similar to the weakly supervised deep detection network (WSDDN) by Bilen \etal\cite{Bilen2016} in network architecture. The major difference is that we replace the detection branch in WSDDN~\cite{Bilen2016} with a elaborately designed region selection module. Concurrent with our work, there are also a series of variants of WSDDN~\cite{Teh2016,Vadim16}. We compare these approaches in Table~\ref{table:waddn_variant}.

\begin{table}[!t]
\centering
    \small
    \caption{Quantitative comparison between our model and other WSDDN variants on the PASCAL VOC 2007 test set.}     
    \begin{tabular}{l l c}
    \hline

    &{\bfseries Method} & {\bfseries mAP } \\

    \hline

     (a) &WSDDN~\cite{Bilen2016} & 31.5 \\
    
     (b) &WSDDN + Sc~\cite{Bilen2016}   & 33.4  \\
    
     (c) &WSDDN + Sc + Reg~\cite{Bilen2016} &34.5 \\

     (d) &AttentionNet~\cite{Teh2016} & 34.5 \\

     (e) &ContextLocNet~\cite{Vadim16}  & 36.3 \\

     (f) &Ours & \textbf{37.4} \\

    \hline
    \end{tabular}
    \label{table:waddn_variant}
\end{table}

As presented in Table~\ref{table:waddn_variant}, our model improves significantly (+5.9\%) over the WSDDN baseline~\cite{Bilen2016} (row (a)), and also outperforms other WSDDN variants. In~\cite{Bilen2016} Bilen \etal extent WSDDN with objectness scaling (row (b)) and a new regularization term (row (c)), boosting their result to 34.5\% in mAP (without model ensemble). Our method still outperforms such variants. In addition, objectness scaling (Sc) cannot be applied to proposal methods which provide no objectness scores for each region (\eg Selective Search~\cite{Uijlings2013}). In contrast, our proposed region selection module does not require extra objectness scores. Our model also outperforms recent variants AttentionNet~\cite{Teh2016} (row (d)) and ContexlocNet~\cite{Vadim16} (row (e)) for object localization under weak supervision. 

\subsection{Compare with state-of-the-art}
We also compare the detection results of our method with recent state-of-the-art WSD methods, including MIL-based methods and other CNN-based models. For a fair comparison, we also re-implement WSDDN~\cite{Bilen2016} as the baseline by removing the region selection module. 

\begin{table*}[!th]
\centering
    \footnotesize
    \setlength\tabcolsep{1.6pt}
    \caption{Quantitative comparison in terms of correct localization (CorLoc) on the PASCAL VOC 2007 trainval set.}     
    \begin{tabular}{l|C C C C C C C C C C C C C C C C C C C C | c}

    \hline

    {\bfseries Method} & {\bfseries aero } & {\bfseries bike } & {\bfseries bird } & {\bfseries boat }&
    {\bfseries bottle } & {\bfseries bus }& {\bfseries car } & {\bfseries cat } & {\bfseries chair } &
    {\bfseries cow } & {\bfseries table } & {\bfseries dog }& {\bfseries horse } & {\bfseries mbike } & {\bfseries pers } & {\bfseries plant } & {\bfseries sheep } & {\bfseries sofa } & {\bfseries train } &
    {\bfseries tv }& {\bfseries mean }\\

    \hline
    Bilen~\cite{Bilen2015}  & 66.4 & 59.3 &42.7 & 20.4 & 21.3 &63.4 &74.3 &\tb{59.6} &21.1&58.2&14.0 &
    38.5 & 49.5 & 60.0 & 19.8 & 39.2 & 41.7 & 30.1 & 50.2 & 44.1 & 43.7 \\

    Cinbis~\cite{Cinbis2016}  & 65.3 & 55.0 &52.4 & \tb{48.3} & 18.2 &66.4 &77.8 &35.6 &\tb{26.5} & 67.0 &46.9 &
    48.4 & \tb{70.5} & 69.1 & 35.2 & 35.2 & 69.6 & 43.4 & 64.6 & 43.7 & 52.0 \\

    Wang~\cite{Wang2015}  & 80.1 & 63.9 &51.5 & 14.9 & 21.0 &55.7 &74.2 &43.5 &26.2 & 53.4 &16.3&
    56.7 & 58.3 & 69.5 & 14.1 & 38.3 & 58.8 & 47.2 & 49.1 & 60.9 & 48.5 \\

    Ren~\cite{Ren2016}  & 79.2 & 56.9 &46.0 & 12.2 & 15.7 &58.4 &71.4 &48.6 &7.2 & \tb{69.9} &16.7 &
    47.4 & 44.2 & 75.5 & \tb{41.2} & 39.6 & 47.4 & 32.3 & 49.8 & 18.6 & 43.9 \\

    Bilen~\cite{Bilen2016}  & 68.5 & 67.5 &56.7 & 34.3 & 32.8 &69.9 &75.0 &45.7 &17.1&68.1&30.5 &
    40.6 & 67.2 & 82.9 & 28.8 & 43.7 & \tb{71.9} & \tb{62.0} & 62.8 & 58.2 & 54.2 \\

    Kantorov~\cite{Vadim16} & \tb{83.3}&68.6&54.7&23.4&18.3&\tb{73.6}&74.1&54.1&8.6&65.1&47.1&\tb{59.5}&67.0&83.5&35.3&39.9&67.0&49.7&63.5&65.2&55.1 \\

    Bency~\cite{Bency2016} & -- & -- & -- & -- & -- & -- & -- & -- & -- & -- & -- &
    -- & -- & -- & -- & -- & -- & -- & -- & -- & 46.8 \\

    \hline
    Baseline  & 69.6 & 68.6 & 58.9 & 26.1 & 40.8 & 68.5 & 70.2 & 36.0 & 11.8 & 58.8 & 36.5&
    41.6 & 66.0 & 82.3 & 17.0 & 46.1 & 56.0 & 39.2 & 70.0 & 67.7 & 51.6 \\

    Ours   & 76.3 & \tb{72.2} &\tb{61.3} &44.1 &\tb{41.2} &70.6 &\tb{78.1} &53.8 & 12.4& 60.3 &\tb{55.5} &
    51.4 & 70.1 & \tb{86.7} & 25.7 & \tb{46.2} & 59.8 & 40.3 & \tb{70.7} & \tb{68.8}& \tb{57.3} \\

    \hline
    \end{tabular}
    \label{table:1}
    \vspace{2mm}
\end{table*}
\begin{table*}[th]
\centering
    \footnotesize
    \setlength\tabcolsep{1.6pt}
    \caption{Quantitative comparison in terms of detection average precision (AP) on the PASCAL VOC 2007 test set.}     
    \begin{tabular}{l|C C C C C C C C C C C C C C C C C C C C | c}

    \hline

    {\bfseries Method} & {\bfseries aero } & {\bfseries bike } & {\bfseries bird } & {\bfseries boat }&
    {\bfseries bottle } & {\bfseries bus }& {\bfseries car } & {\bfseries cat } & {\bfseries chair } &
    {\bfseries cow } & {\bfseries table } & {\bfseries dog }& {\bfseries horse } & {\bfseries mbike } & {\bfseries pers } & {\bfseries plant } & {\bfseries sheep } & {\bfseries sofa } & {\bfseries train } &
    {\bfseries tv } & {\bfseries mean }\\

    \hline
    Bilen \cite{Bilen2015}  & 46.2 & 46.9 &24.1 & 16.4 & 12.2 &42.2 &47.1 &\tb{35.2} &7.8& 28.3 & 12.7 &
    21.5 & 30.1 & 42.4 & 7.8 & 20.0 & 26.8 & 20.8 & 35.8 & 29.6 & 27.7 \\

    Cinbis \cite{Cinbis2016}  & 39.3 & 43.0 &28.8 & 20.4 & 8.0 &45.5 &47.9 &22.1 &8.4 & 33.5 &23.6 &
    29.2 & 38.5 & 47.9 & 20.3 & 20.0 & 35.8 & 30.8 & 41.0 & 20.1 & 30.2 \\

    Wang \cite{Wang2015}  & 48.9 & 42.3 &26.1 & 11.3 & 11.9 &41.3 &40.9 &34.7 &\tb{10.8} & 34.7 &18.8&
    34.4 & 35.4 & 52.7 & 19.1 & 17.4 & 35.9 & 33.3 & 34.8 & 46.5 & 31.6 \\

    Ren \cite{Ren2016}  & 41.3 & 39.7 &22.1 & 9.5 & 3.9 &41.0 &45.0 &19.1 &1.0 & 34.0 &16.0&
    21.3 & 32.5 & 43.4 & \tb{21.9} & 19.7 & 21.5 & 22.3 & 36.0 & 18.0 & 25.4 \\

    
    Bilen \cite{Bilen2016}  & 42.9 & 56.0 &32.0 & 17.6 & 10.2 &61.8 &50.2 &29.0 &3.8& 36.2 & 18.5 &
    31.1 & 45.8 & 54.5 & 10.2 & 15.4 & \tb{36.3} & 45.2 & 50.1 & 43.8 & 34.5 \\

    Kantorov~\cite{Vadim16} & \tb{57.1}&52.0&31.5&7.6&11.5&55.0&53.1&34.1&1.7&33.1&\tb{49.2}&\tb{42.0}&47.3&56.6&15.3&12.8&24.8&\tb{48.9}&44.4&47.8&36.3 \\

    Bency \cite{Bency2016} & -- & -- & -- & -- & -- & -- & -- & -- & -- & -- & -- &
    -- & -- & -- & -- & -- & -- & -- & -- & -- & 25.7 \\


    \hline
    Baseline &42.1&50.8&29.7&18.5&14.7&56.8&48.3&14.2&3.1&32.3&24.2&24.7&48.9&53.8&6.6&19.9&24.8&24.2&54.5&44.5&31.8 \\

    Ours    &50.1&\tb{56.1}&\tb{33.4}&\tb{21.1}&\tb{17.8}&\tb{62.0}&\tb{54.2}&34.2&3.1&\tb{37.1}&38.5&32.5&\tb{52.9}&\tb{57.0}&5.3&\tb{21.4}&30.0&31.6&\tb{57.4}&\tb{50.2}&\textbf{37.4} \\
    \hline
    \end{tabular}
    \label{table:2}
    \vspace{2mm}
\end{table*}

Table~\ref{table:1} shows performance comparison in terms of CorLoc~\cite{Deselaers2012} on the PASCAL VOC 2007 trainval set. Our method achieves 57.3\% of average CorLoc for all the 20 categories, outperforming all the alternatives. It indicates that our model achieves the best performance of localizing true positive regions in the training set, which in another way verifies the effectiveness of the proposed region selection module. We also present the average precision (AP) on the PASCAL VOC 2007 test set in Table~\ref{table:2}. Our model achieves 37.4\% on mAP, which outperforms all the competitors.

With optimized region selection module, the learned object detectors outperform the baseline on 19 out of 20 categories in terms of both localization and detection. Compared to the best MIL-based approaches~\cite{Wang2015,Cinbis2016}, we achieve significant improvements by 5.3\% in Corloc, and 5.8\% in mAP. For most MIL-based methods, positive training regions are selected from a noisy candidate set. We believe the large amount of background clutters within the candidate set hurt the precision of the positive training set, which leads to large object-background ambiguity. In contrast, our model progressively remove the noisy backgrounds from the positive set. As a result, the precision of the collected positive set improves and high recall rate is also retained, which makes the learned object detectors discriminative. The class-specific negative set also contributes to the performance boost. Compared with CNN-based models, our model outperforms~\cite{Bency2016} by 10.5\% in CorLoc and 11.7\% in mAP. We notice that in~\cite{Bency2016}, class activation maps are learned from the entire image. The noisy background regions inevitably affect the accuracy of the learned activation maps, and therefore may hurt detection performance. In comparison, our proposed region selection module effectively alleviates the impact from background clutters, and therefore achieves better detection performance. Visual detection results on the PASCAL VOC 2007 test set are shown in the Appendix.

We further perform experiments on the PASCAL VOC 2010 and VOC 2012 dataset. As presented in Table~\ref{table:3}, our method achieves mAP of 36.0\% on VOC 2010 and 33.6\% on VOC 2012, significantly outperforms the state-of-the-art methods.

\begin{table}[!t]
\centering
    \small
    \setlength\tabcolsep{10pt}
    \caption{Quantitative comparison in terms of detection average precision (AP) on the PASCAL VOC 2010 and VOC 2012.}     
    \begin{tabular}{l c c}
    \hline
    \bfseries{Method} & {\bfseries VOC 2010} & {\bfseries VOC 2012} \\
    \hline

     Oquab~\cite{Oquab2015} & - & 11.7 \\

     Cinbis \cite{Cinbis2016} & 27.4 & - \\
    
     Ren~\cite{Ren2016} & - & 23.8  \\
    
     Bency~\cite{Bency2016} & - &26.5 \\

     Baseline & 29.8 & 29.6 \\

     Ours  & \textbf{36.0} & \textbf{33.6} \\

    \hline
    \end{tabular}
    \label{table:3}
    \vspace{2mm}
\end{table}

\subsection{Visualization analysis}
\begin{figure*}[!h]
\centering
\epsfig{file=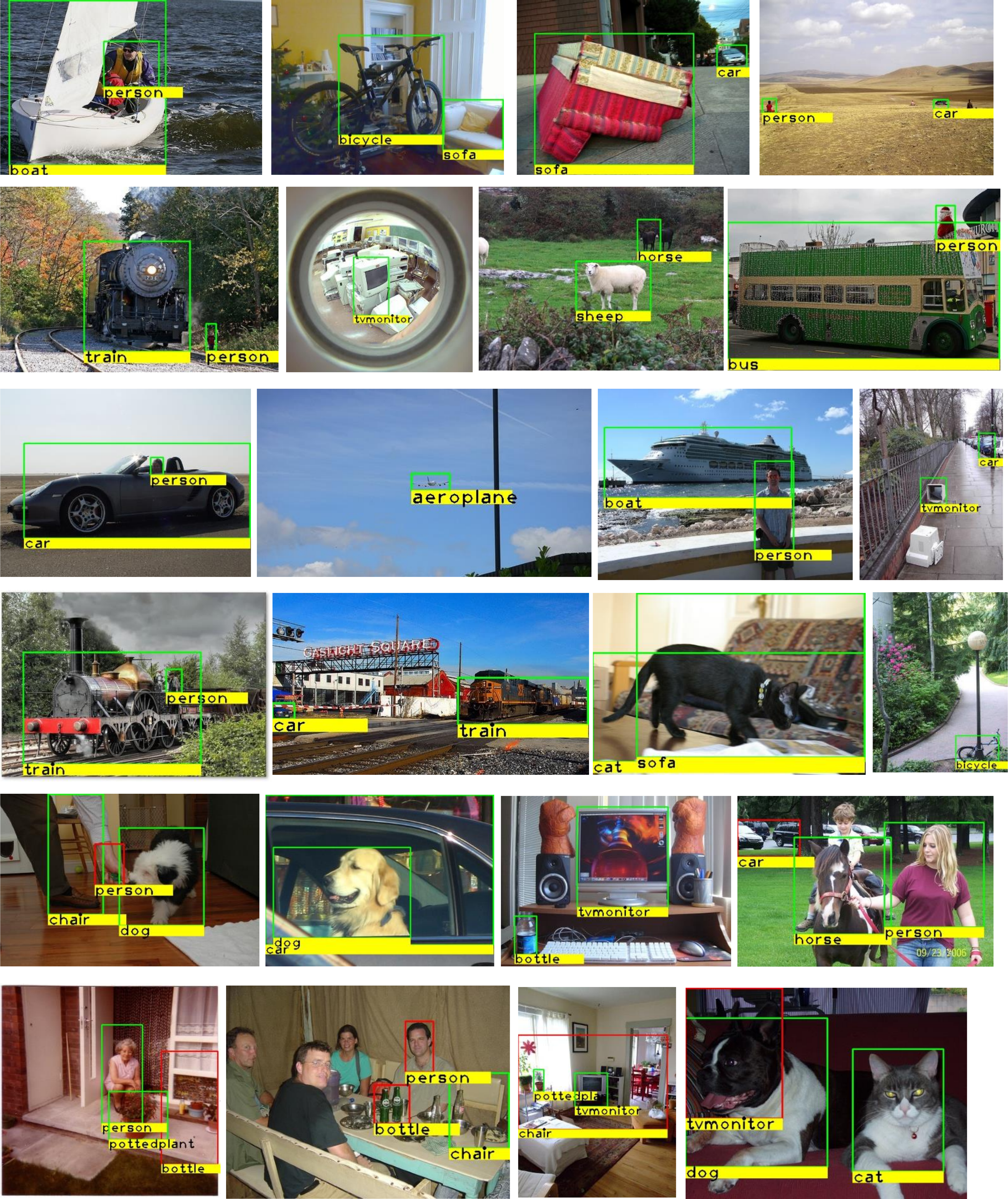, width=\linewidth}
\caption{Sample detection results. For all the examples, we only show the top one detected region for each category. Green boxes indicate correct detections, red boxes indicate false detections.}
\label{fig:6}
\end{figure*}

Visual detection results on the PASCAL VOC 2007 test set are shown in Figure~\ref{fig:6}. For each category, the box with the highest prediction score is drawn. From these examples, it can be seen that our proposed method is able to localize objects subject to great variability in scale and appearance, and even some of the small target objects are accurately discovered. Many heavily occluded objects are also successfully localized. However, several bad cases still exist. One problem is inaccurate box prediction, where only part of the true object is captured or too much background is covered. For example, when detecting a person, the bounding box is sometimes drawn on person's upper body. Hence, the actual performance for person detection is not desirable. Another common problem is the confusion of several visual similar classes. We believe that improving the quality of positive mining by considering both inter-class and intra-class metrics in our formulation is one way to handle these problems.

\section{Acknowledgment}
This work is supported by Chinese National Natural Science Foundation (61471049, 61372169, 61532018) and China Scholarship Council.

\section{Conclusion}
We propose an optimized region selection strategy for weakly supervised detection. This method collects purified positive training regions by progressively removing easy background clutters, which improves the precision of the positive set and retains high diversity of the training samples as well. This approach also selects discriminative negative samples by mining class-specific hard negatives. The region selection module is combined with the learning of object detectors so that both parts can be jointly optimized during training. 
We extensively evaluate the detection performance on the PASCAL VOC 2007, VOC 2010 and 2012 datasets. Experimental results demonstrate the region selection strategy effectively improves weakly supervised visual learning, and could become common practice for weakly supervised detection.


{\small
\bibliographystyle{ieee}
\bibliography{wenhui}
}

\end{document}